\documentclass[11pt]{article}

\usepackage[preprint]{acl}

\usepackage{times}
\usepackage{latexsym}
\usepackage{algorithm}
\usepackage{algorithmic}
\usepackage{amsmath}
\usepackage{amssymb}
\usepackage[most,breakable]{tcolorbox}
\usepackage{listings}
\usepackage[T1]{fontenc}

\usepackage[utf8]{inputenc}

\usepackage{microtype}

\usepackage{inconsolata}

\usepackage{graphicx}

\usepackage{booktabs}
\usepackage{multirow}

\title{Alleviating Choice Supportive Bias in LLM with Reasoning Dependency Generation}

\author{
  Nan Zhuang$^{*}$ \quad
  Wenshuo Wang$^{1*}$ \quad
  Lekai Qian$^{1*}$ \quad
  Yuxiao Wang$^{1}$ \quad
  Boyu Cao$^{1}$ \quad
  Qi Liu$^{1\dagger}$ \\[1ex]
  $^1$School of Future Technology, South China University of Technology, Guangzhou, PRC \\[0.5ex]
  \texttt{@gmail.com,\{202364870251,ftqlk,202411094427,drliuqi\}@scut.edu.cn} \\
  $^*$Equal contribution \quad $^\dagger$Corresponding author
}

\begin{document}
\maketitle
\begin{abstract}
Recent studies have demonstrated that some Large Language Models exhibit choice-supportive bias (CSB) when performing evaluations, systematically favoring their chosen options and potentially compromising the objectivity of AI-assisted decision making. While existing debiasing approaches primarily target demographic and social biases, methods for addressing cognitive biases in LLMs remain largely unexplored. In this work, we present the first solution to address CSB through Reasoning Dependency Generation (RDG), a novel framework for generating unbiased reasoning data to mitigate choice-supportive bias through fine-tuning. RDG automatically constructs balanced reasoning QA pairs, explicitly (un)modeling the dependencies between choices, evidences, and justifications. Our approach is able to generate a large-scale dataset of QA pairs across domains, incorporating Contextual Dependency Data and Dependency Decouple Data. Experiments show that LLMs fine-tuned on RDG-generated data demonstrate a 81.5\% improvement in memory-based experiments and 94.3\% improvement in the evaluation-based experiment, while maintaining similar performance on standard BBQ benchmarks. This work pioneers an approach for addressing cognitive biases in LLMs and contributes to the development of more reliable AI-assisted decision support systems.
\end{abstract}

\begin{figure*}[]
\centerline{\includegraphics[width=0.8\textwidth]{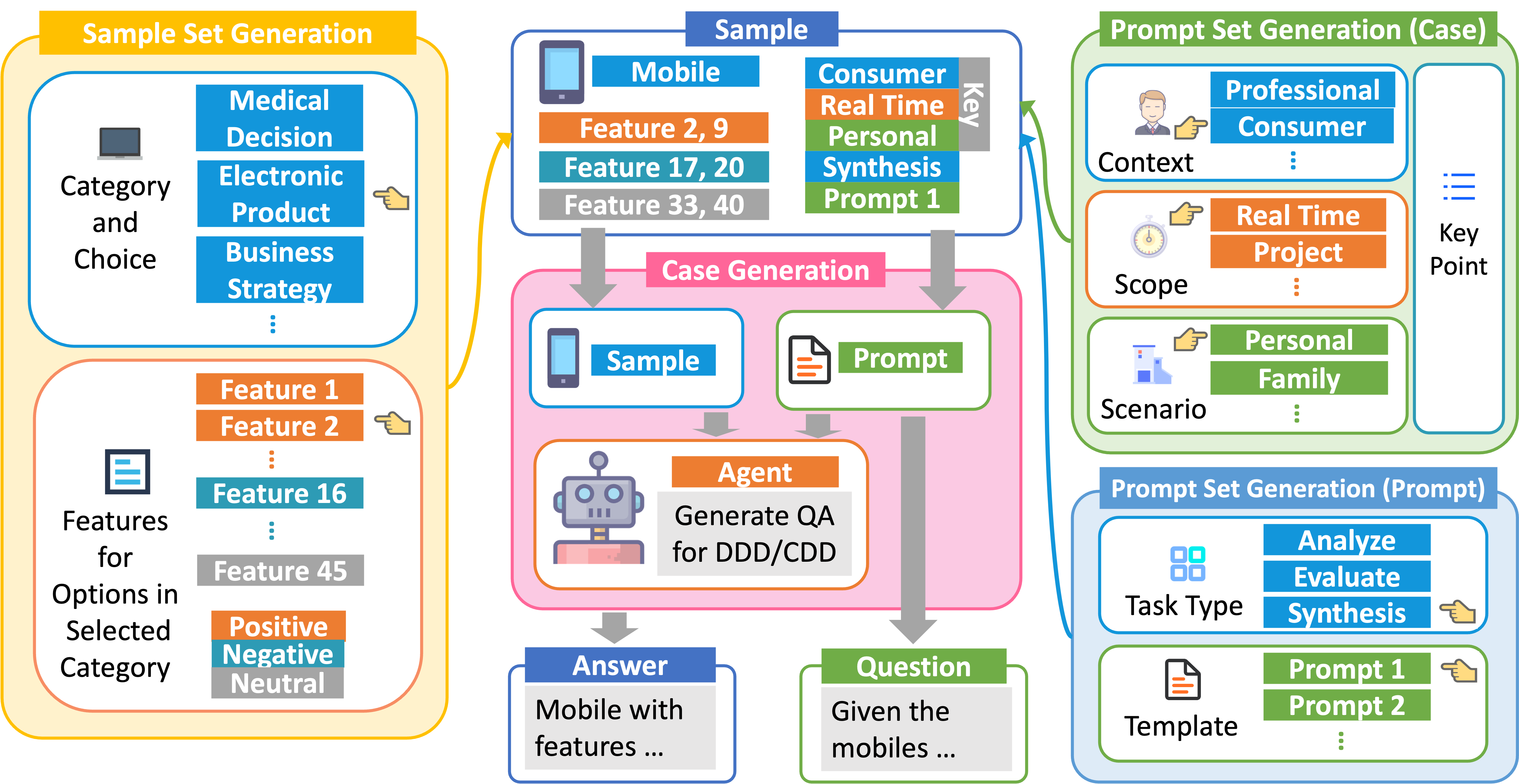}} 
\caption{The overview for the proposed RDG method.}
\label{fig:proposed_method_illustration}
\end{figure*}

\section{Introduction}
The integration of large language models (LLMs) into high-stakes decision-making - from guiding clinical diagnostics \cite{kaur2024llm} to optimizing financial portfolios \cite{park2024enhancing} - demands objective reasoning integrity. While extensive research has addressed demographic and social biases in LLMs (e.g., gender stereotypes, racial prejudices), a critical category remains largely unexplored: cognitive biases - systematic deviations in reasoning processes that affect how decisions are justified and evaluated \cite{echterhoff2024cognitive}. Unlike demographic biases that manifest as unfair treatment of specific groups, cognitive biases distort the fundamental logic chains LLMs use to support their choices. Among these, choice-supportive bias (CSB) - the tendency to retroactively attribute more positive qualities to chosen options and more negative qualities to rejected ones \cite{mather2000choice} - poses unique risks in AI-assisted decision-making. Consider a medical AI that recommends Treatment A over Treatment B: with CSB, the model might subsequently overemphasize A's benefits (``minimal side effects") while exaggerating B's drawbacks (``high complication risk"), even when objective evidence suggests comparable outcomes. Such post-hoc rationalization can entrench suboptimal decisions and prevent proper re-evaluation when new information emerges.

CSB fundamentally differs from the demographic and social biases that dominate current debiasing research. Traditional biases target fixed attributes - a gender bias consistently favors males over females \cite{buolamwini2018gender}, a racial bias systematically disadvantages certain ethnicities. In contrast, CSB is context-dependent: it favors whatever option was chosen, regardless of its inherent qualities. This dynamic nature means CSB may not be addressed well by existing debiasing methods, which are designed to correct static correlations between protected attributes and outcomes. When standard debiasing techniques balance demographic representation or filter biased training examples, they may miss CSB because the bias target shifts with each decision context. Recent empirical work confirms this challenge: \cite{zhuang2025llm} demonstrated that even commercial LLMs exhibit significant CSB, systematically inflating positive assessments of chosen options significantly across diverse domains. As debiasing research for cognitive biases remains nascent compared to demographic debiasing, new approaches are urgently needed to ensure AI systems can maintain objective reasoning throughout decision chains.

Reasoning dependency describes the pattern of how LLMs should depend on different information sources when constructing evaluations and explanations. Proper reasoning dependency requires models to primarily attend to factual context while maintaining independence from choice status during evaluation. However, current LLMs may exhibit faulty dependencies: over-attending to whether an option was chosen while under-utilizing available facts. This distorted attention pattern propagates through chain-of-thought reasoning tasks, where objective evaluation is crucial for reliable outcomes. While existing debiasing methods focus on surface-level corrections, addressing CSB requires tuning the fundamental attention to ensure unbiased reasoning trajectories.

Therefore, we propose Reasoning Dependency Generation (RDG), the first framework that mitigates CSB by systematically tuning how LLMs associate between decisions, facts, and justifications. RDG employs a three-stage pipeline: first, generating domain-agnostic sample set using controlled templates that isolate choice-rationale dependencies and provide simulated factual information; second, producing a prompt set, with which LLMs perform varied tasks that does not need to attend their choice; and third, sampling from the sets and synthesizing specific and balanced QA examples. Crucially, RDG introduces counterfactual prompting, where models must defend unchosen (rejected) options using identical evidence, disrupting entrenched preference-justification links. The effectiveness of this relatively small sample size highlights that diversity of reasoning patterns, rather than sheer quantity of examples, is the key factor in addressing choice-supportive bias. By fine-tuning on just 2500 randomly sampled pairs, RDG compels models to internalize symmetry between chosen and rejected options, addressing a flaw inherent in conventional fine-tuning data that may prioritize outcome over reasoning integrity. 

Experiments demonstrate RDG's efficacy: it reduces CSB-driven attribution errors by 83.7\% in memory tasks (measured by average reduction in positive/negative feature attribution differences) and decreases tendency score disparities by 94.7\% in evaluations (measured by average scoring bias) while maintaining similar performance on BBQ \cite{parrish2021bbq}. These advances position RDG as a critical step toward LLMs that combine human-like reasoning flexibility with algorithmic impartiality on CSB.


\section{Related Works}\label{sec:2}

\subsection{Choice-supportive Bias in Human and AI Systems}
Choice-supportive bias (CSB) - the systematic attribution of positive traits to chosen options - originates in human cognitive architecture. Cognitive biases are predictable patterns of deviation from rationality in judgment, with CSB being a specific variant that affects decision evaluation. Foundational studies demonstrated CSB's persistence across various tasks \cite{mather2000choice,henkel2007memory,lind2017choice}, with some research linking it to escalated commitment \cite{zorn2020impact}.

In LLMs, while cognitive biases like framing effects and confirmation bias have been studied \cite{suri2024large,shi2024argumentative}, CSB was empirically identified through two paradigms \cite{zhuang2025llm}: 1) memory-driven contextual hallucination favoring chosen options, and 2) evaluation-based distortion of neutral features. This bias may propagate dangerously in sequential decision systems-clinical diagnosis agents \cite{kaur2024llm} and financial models \cite{park2024enhancing} increasingly rationalize suboptimal choices through self-reinforcing logic.

Unlike widely-researched demographic biases \cite{buolamwini2018gender}, CSB corrupts reasoning processes rather than surface outputs: agents construct plausible but skewed causal narratives \cite{wang2023can}. Traditional debiasing approaches may fail to address this process-level distortion, creating critical reliability gaps in LLM-based decision architectures. Notably, no methods have directly targeted CSB in LLMs prior to this work.

\subsection{Debiasing Approaches in Language Models}

Current debiasing methods primarily target static demographic biases rather than dynamic cognitive biases like CSB. Pre-processing techniques filter data \cite{dhamala2021bold} or generate moral exemplars \cite{sun2022moraldial}, while in-training methods use debiasing adapters \cite{lauscher2021sustainable}. Post-processing approaches include self-debiasing prompts \cite{schick2021self} and output rewriting \cite{tokpo2022text}, though word-swapping risks factual inconsistencies \cite{kumar2022language} and decoding constraints prioritize plausibility over logical consistency.

These methods may not work well for CSB because they correct static correlations between protected attributes and outcomes, not dynamic reasoning dependencies. CSB manifests as distorted attention patterns where models over-rely on choice status while under-utilizing factual context - a fundamentally different problem than demographic bias. Data augmentation oversimplifies these dynamics \cite{devinney2022theories}, while post-hoc editing leaves biased reasoning chains intact. CSB's context-dependent nature, where bias targets shift with each decision, renders static correction ineffective.

Prompt-based methods offer potential for cognitive bias mitigation by leveraging LLMs' capabilities directly. These include chain-of-thought \cite{kamruzzaman2024prompting}, self-debiasing \cite{schick2021self}, prompting strategies \cite{si2022prompting}, and counterfactual debiasing \cite{li2024prompting}. While promising for their ability to modify reasoning processes, these methods were developed for demographic biases, with unverified effectiveness for CSB, motivating our comparative evaluation.

To our knowledge, no methods directly address CSB in LLMs. RDG fills this gap through structured reasoning dependency generation, dynamically varying choice-evidence-justification connections. Unlike static moral exemplars \cite{sun2022moraldial}, RDG forces models to disentangle preference from validity during objective evaluation, directly targeting choice-supportive bias's root cause.

\section{Proposed Method}

\subsection{Overview}
Choice-supportive bias in LLMs manifests as problematic attention patterns during evaluation: models over-rely on an option's chosen/unchosen status while under-utilizing factual evidence. This bias operates through learned dependencies—the implicit associations between information types that guide reasoning. In unbiased reasoning, evaluations should depend primarily on factual features, not choice status. However, LLMs learn spurious correlations from training data where positive justifications consistently accompany chosen options, creating incorrect dependencies between choice status and evaluation valence. This insight motivates a more fundamental solution: restructuring the information dependencies themselves through targeted training data that explicitly varies relationships between choices, facts, and evaluations. By teaching models which information should and should not influence their reasoning, we can address CSB at its root.

We propose Reasoning Dependency Generation (RDG), a framework that creates training data to recalibrate how LLMs process decision-related information (Figure \ref{fig:proposed_method_illustration}). RDG employs two complementary strategies: Contextual Dependency Data (CDD) strengthens appropriate dependencies by creating scenarios where correct evaluations require attention to facts rather than choice labels; Dependency Decoupling Data (DDD) breaks inappropriate dependencies by requiring identical features to be evaluated as both chosen and unchosen options. Both process rely on the same generated background dataset. Through systematic generation of such examples, RDG restructures the learned dynamic associations that cause choice-supportive bias.

\subsection{Background Dataset Generation}
We propose a two-part dataset architecture consisting of a Sample Set for core decision elements and a Prompt Set for instruction contexts, as illustrated in the left and right sections of Figure \ref{fig:proposed_method_illustration}. This foundation enables generating both Contextual Dependency Data (CDD) and Dependency Decoupling Data (DDD) while maintaining controlled variation across decision scenarios.

The Sample Set employs domain-abstracted categories (e.g., Medical Decision) to prevent models from leveraging pre-trained biases and opinions. Within each category, we synthesize options using controlled attribute generation. For instance, instead of comparing specific products, options might be designated as ``Car A" versus ``Car B," forcing evaluation based purely on provided features rather than prior knowledge. For each option, we generate balanced feature sets through a multi-stage process. Features are then generated using constrained prompting that enforces concrete, or perceived attributes - positive features like ``Enables 40\% efficiency improvement", negative features like ``Requires 3-month initialization period", and neutral features like ``Employs modular architecture". Each feature undergoes classification by two independent LLM agent evaluators to confirm its positive/negative/neutral designation, with the confidence averaged and assigned to the choice for later use.

The Prompt Set introduces structured evaluation contexts through systematic variation of three key dimensions: evaluator context (e.g., professional, consumer, personal), temporal scope (real-time, project-scope, long-term impacts), and usage scenarios (personal, family, company). Each dimension supports 5 distinct values, generating sufficient unique evaluation contexts. For each context combination, we employ a two-stage CoT process to generate assessment criteria (key points): first, a LLM agent identifies relevant evaluation factors; then, a separate agent uses these factors for generating one paragraph describing the criteria. For example, the combination {Expert evaluator, Long-term impact, Resource-constrained environment} generates criteria focused on sustainability, optimization potential, and maintenance requirements - factors that apply equally to chosen and unchosen options. Drawing from cognitive psychology research, our prompt templates incorporate three higher-order cognitive task types identified by Miri et al. \cite{miri2007purposely}: analysis (breaking down complex features), evaluation (assessing relative importance), and synthesis (integrating multiple factors). These cognitive frameworks ensure prompts engage comprehensive reasoning rather than superficial comparisons. To prevent exploitation of prompt patterns, we implement template variation with controlled paraphrasing that preserves semantic equivalence while altering surface structure. This includes permuting question framing (e.g., ``Evaluate X considering Y" vs. ``Given Y, assess X"), context presentation order, and response elicitation methods while maintaining consistent underlying tasks. Each template variant undergoes validation to ensure it elicits equivalent reasoning patterns across different models.

\subsection{Contextual Dependency Data (CDD) Generation with Agents}

CDD generation (Algorithm \ref{alg:CDD} in the Appendix) aims to create training examples that explicitly model correct factual reasoning dependencies, helping LLMs learn to prioritize factual context over choice information in their evaluations. The generated QA pairs focus on two key patterns of objective reasoning. First, pairs require the model to assess preferred options using only provided facts. Second, feature attribution pairs test the model's ability to correctly match characteristics to options without choice-based bias. This directly addresses a core mechanism of choice-supportive bias in LLM - the tendency to under-utilize available factual context while reasoning. 

The algorithm begins by sampling a category and options (line 3) from the Sample Set. We then sample feature sets for options through controlled sampling. Rather than using fixed positive-negative ratios, we employ a beta distribution ($\alpha$=2, $\beta$=2) to determine positive/negative feature proportions during feature sampling (line 4), creating natural variation in complexity while maintaining overall balance, while preventing models from learning simplistic heuristics based on feature counts. Features undergo validation through conflict checking (lines 6-9) using LLM-based semantic similarity checking to ensure no contradictions within options (threshold 0.6) and no excessive similarity between options (threshold 0.75). These thresholds were empirically determined to maximize feature distinctiveness while maintaining realistic scenario coherence.

For case generation, phase creates two QA pairs for each scenario - an option analysis phase (generate analysis and selection, line 12) and a fact-related answer phase (line 13-15). For the option analysis phase, we combine validated features with templates from our Prompt Set, selecting from our three cognitive task types (analysis, evaluation, synthesis) to create diverse reasoning challenges. Each template incorporates the sampled evaluation context (e.g., {Professional evaluator, Project scope, Team environment}) and its corresponding assessment criteria. To generate reference answers, we use the target model itself with controlled temperature settings ($T=0$), preserving its base reasoning capabilities while focusing on dependency restructuring. This approach ensures that generated answers maintain original language patterns while explicitly demonstrating desired reasoning dependency.

The fact-related answer phase asks question directly about given facts. It extends each option by introducing new distractor features. These distractors are sampled and validated in the same way as previously mentioned to ensure distinctiveness. Rather than using binary correctness labels, we assign graduated confidence scores using a uniform distribution (line 14), 0.7-0.9 for positive feature associations and 0.6-0.8 for negative ones. This subtle difference in confidence ranges reflects the empirically observed tendency for positive attributes to be more distinctly memorable in human studies, while the uniform distribution prevents the model from learning to exploit fixed confidence patterns (e.g., 1.0).

\subsection{Dependency Decouple Data (DDD) Generation}
The Dependency Decouple Data (DDD) generation process (Algorithm \ref{alg:DDD} in the Appedix) specifically targets the tendency of LLMs to link choice status with objective evaluation. Where CDD strengthens factual context dependencies, DDD explicitly breaks learned associations between an option's chosen/unchosen status and how its characteristics are evaluated. In the QA pair, for each feature, the model is prompted to evaluate it both as part of a chosen and an unchosen option. For instance, when assessing "modular architecture", one question analyzes it in chosen Option A, while its pair evaluates the same feature in chosen Option B. This structure forces the model to maintain consistent reasoning dependencies based on the feature's inherent characteristics rather than choice status. This complementary approach creates training examples that force models to assess features independently of choice information.

DDD employs a balanced feature structure to enable rigorous pattern learning. For each scenario, we sample exactly \textit{N} positive, \textit{N} negative, and \textit{N} neutral features per option, maintaining strict distributional equality. Feature validation follows patterns similar to CDD. The algorithm creates paired QA examples (lines 11-16) by varying how identical features are assessed. For each scenario, we first generate a standard evaluation Q for a randomly chosen option (lines 11-12), using our sampled features and context. We then create complementary evaluation for the unchosen option (lines 14-15) by randomly reassigning choice status and reordering options while maintaining the same underlying features. This paired structure forces models to consider identical attributes both as chosen and unchosen characteristics. The evaluation answers (line 13) incorporate context variables (task type, temporal scope, usage scenario) to ensure evaluations remain realistic while focusing purely on feature assessment rather than choice justification. A field of objectivity was assigned to the evaluation justification (in genEvalA), encouraging the LLM to be objective by classifying the feature itself into positive, negative, and neutral. The resulting dataset encourages learned dependencies between choice status and feature evaluation, directly supporting our framework's goal of comprehensive bias mitigation through reasoning restructuring.

\begin{table*}[htbp]

\centering
\caption{Memory-based Experiment - Seen Features Difference Results with ANOVA C(option) F}
\resizebox{\textwidth}{!}{
\setlength{\tabcolsep}{3pt} 
\begin{tabular}{l|cccc|cccc|cccc|cccc}
\toprule
\multirow{2}{*}{Method} & \multicolumn{4}{c|}{GPT-4o-mini-2024-07-18} & \multicolumn{4}{c|}{GPT-4o-2024-08-06} & \multicolumn{4}{c|}{qwen3-235b-a22b} & \multicolumn{4}{c}{deepseek-r1} \\
 & P-Diff & N-Diff & SE & C(opt)F & P-Diff & N-Diff & SE & C(opt)F & P-Diff & N-Diff & SE & C(opt)F & P-Diff & N-Diff & SE & C(opt)F \\
\midrule
Baseline & .321 & .096 & .065 & 506.5 & .120 & .000 & .050 & 151.0 & .070 & .002 & .074 & 125.6 & .057 & -.041 & .083 & 287.3 \\
\textbf{RDG} & \textbf{-.068} & \textbf{.000} & \textbf{.045} & \textbf{4.3} & \textbf{.010} & \textbf{.000} & \textbf{.035} & \textbf{1.9} & \textbf{.028} & \textbf{-.001} & \textbf{.035} & \textbf{1.1} & \textbf{.006} & \textbf{-.003} & \textbf{.040} & \textbf{2.1} \\
DDD & .358 & .115 & .068 & 558.4 & .138 & .012 & .055 & 168.5 & .082 & .008 & .078 & 142.8 & .068 & -.048 & .092 & 312.6 \\
CDD & .145 & .042 & .055 & 85.7 & .054 & -.002 & .042 & 32.6 & .035 & .001 & .058 & 25.3 & .023 & -.018 & .065 & 45.2 \\
CoT & .162 & .045 & .052 & 12.5 & .088 & .000 & .048 & 8.8 & .282 & .158 & .051 & 29.6 & .030 & -.053 & .045 & 6.9 \\
Self-debias & .177 & .060 & .065 & 15.3 & .132 & .079 & .055 & 11.0 & .177 & .062 & .049 & 18.8 & .100 & -.008 & .062 & 9.2 \\
Explicit & .123 & .022 & .048 & 9.8 & .072 & .000 & .045 & 6.6 & .070 & .002 & .050 & 7.8 & .081 & .003 & .055 & 5.4 \\
Counterfact. & .217 & .074 & .072 & 18.7 & .156 & .080 & .058 & 12.5 & .397 & .245 & .051 & 31.5 & .488 & .499 & .057 & 45.8 \\
\bottomrule
\end{tabular}
}
\\[0.5em]
\footnotesize
\textbf{Note:} The closer diff value to zero, the lower SE, the better. For C(option)F, lower values indicate better performance.
\label{tab:mem}
\end{table*}

\section{Experiment}
This section presents the evaluations of the proposed bias mitigation approach on two experiments. While we considered comparing against traditional debiasing methods, they primarily target static correlational biases and are not applicable as baselines for addressing CSB (As mentioned in the Related Works), so we only compared with prompt-based methods \cite{kamruzzaman2024prompting,schick2021self,si2022prompting,li2024prompting}. These experiments are directly adapted from the comprehensive CSB assessment framework established in previous study \cite{zhuang2025llm}, allowing for consistent comparison between our RDG-mitigated models and previously identified baseline behaviors. We further conducted BBQ and ablation studies to further prove the reliability of our method.

\subsection{Dataset Generation and Fine Tuning}
We selected gpt-4o, gpt-4o-mini, qwen3-235b-a22b, deepseek-r1 as our baseline models. This selection enables direct comparability with previous findings while focusing on a widely deployed model family that includes both open/close source, and serves millions of users daily in decision-support applications.

We generated two large-scale datasets following the RDG framework: Contextual Dependency Data (CDD) and Dependency Decouple Data (DDD). Each dataset contains examples distributed across domains, and was generated with the same baseline model with $T=0$ and \textit{Top\_p}=$1.0$. While the full dataset represents a diverse space of possible reasoning dependencies, we use experiments to demonstrate that randomly sampling just 2,500 examples in total is sufficient for effective bias mitigation. Our training process followed a two-phase schedule optimized for bias mitigation. In the first phase, we fine-tuned the baseline model on 2,000 randomly sampled DDD examples, strengthening foundational attention for factual information. This initial training helps the model learn to evaluate options independently of choice status. The second phase introduced 500 randomly sampled CDD examples, teaching the model to maintain this balanced reasoning while preserving contextual awareness. This sequential approach allows the model to first develop robust debiasing patterns before learning to apply them in context-rich scenarios.

We employed supervised fine-tuning using OpenAI's API service and self-hosted service, with a learning rate multiplier of 0.1 for GPT, lr=5e-6 for Qwen3 and Deepseek-r1, and limited training to 3 epoch to prevent overfitting. The relatively small training set size (2,500 examples total) proved sufficient for achieving significant bias reduction while maintaining model capabilities. We validated the training using a held-out set of 2,000 DDD and 500 CDD examples, shuffled to ensure representative sampling across steps. This training configuration balances the competing demands of bias mitigation and capability preservation, as demonstrated in our experimental results.

\subsection{Memory-based Experimental Settings}  
We replicate the memory-based protocol and metrics from \cite{zhuang2025llm} to evaluate how RDG restructures choice-justification dependencies. The experiment employs the 
paradigm from \cite{fechner1860elemente} using decision scenarios from the original study’s domains. Each scenario presents five options with ten features each. We contrast the original model with our RDG-fine-tuned variant under same conditions. Feature order randomization and distractor insertion follow the original implementation, with 50 trials per model to match the prior work.

The experimental design preserves the two-phase structure while emphasizing reasoning chain analysis. During the selection phase, models receive scenario descriptions, requiring forced choices between options with randomized feature selection and order. The recall phase then queries for which feature belongs to which choice. Responses are constrained to JSON formats to ensure metric compatibility.

We employ two core metrics from the foundational study. The Chosen-Rejected Accuracy Differential quantifies memory fidelity disparities using normalized accuracy differences in positive/negative feature attribution between chosen and rejected options. This is calculated separately for original features (seen) and new distractors (unseen). 
Temperature is fixed at 0 across all trials to avoid stochastic effects, with prompt phrasing standardized to the original study’s exact formulations.

\begin{table*}[h!]

\centering
\caption{Evaluation-based Experiment - Tendency Scores}
\small
\setlength{\tabcolsep}{3pt} 
\begin{tabular}{l|ccc|ccc|ccc|ccc}
\toprule
\multirow{2}{*}{Method} & \multicolumn{3}{c|}{GPT-4o-mini-2024-07-18} & \multicolumn{3}{c|}{GPT-4o-2024-08-06} & \multicolumn{3}{c|}{qwen3-235b-a22b} & \multicolumn{3}{c}{deepseek-r1} \\
 & Chosen & Rejected & p-val & Chosen & Rejected & p-val & Chosen & Rejected & p-val & Chosen & Rejected & p-val \\
\midrule
Baseline & 1.76 & -2.54 & $<$.001 & 2.30 & -3.00 & $<$.001 & 0.23 & -0.23 & .005 & 2.07 & -3.30 & $<$.001 \\
\textbf{RDG} & \textbf{0.13} & \textbf{-0.10} & \textbf{.318} & \textbf{0.08} & \textbf{-0.12} & \textbf{.185} & \textbf{0.02} & \textbf{-0.02} & \textbf{.652} & \textbf{0.11} & \textbf{-0.17} & \textbf{.028} \\
DDD & 0.25 & -0.18 & .003 & 0.32 & -0.28 & $<$.001 & 0.03 & -0.02 & .482 & 0.38 & -0.42 & $<$.001 \\
CDD & 0.95 & -1.25 & $<$.001 & 1.28 & -1.72 & $<$.001 & 0.18 & -0.15 & .042 & 1.45 & -2.05 & $<$.001 \\
CoT & 0.35 & -0.42 & .002 & 0.65 & -0.78 & $<$.001 & 0.00 & -0.01 & .319 & -0.02 & -0.01 & .563 \\
Self-debias & 0.52 & -0.78 & $<$.001 & 0.82 & -1.05 & $<$.001 & -0.27 & 0.06 & .006 & -0.46 & 1.76 & $<$.001 \\
Explicit & 0.28 & -0.35 & .008 & 0.42 & -0.58 & $<$.001 & 0.01 & 0.50 & .066 & -0.17 & -0.19 & .808 \\
Counterfact. & 0.45 & -0.68 & $<$.001 & 0.75 & -0.95 & $<$.001 & 0.02 & 0.23 & .003 & 0.00 & -0.01 & .319 \\
\bottomrule
\end{tabular}
\\[0.5em]
\footnotesize
\textbf{Note:} For chosen and unchosen, the closer to zero, the better; for the p-val, the higher, the better. See reference for formulas used for calculations.
\label{tab:eval}
\end{table*}

\subsection{Memory-based Experimental Results}
Our memory-based experiments demonstrate RDG's effectiveness in restructuring how models process and utilize choice-related information. Table \ref{tab:mem} and \ref{tab:general} were aggregated from tested accuracy data (see Appendix), showing that the RDG-tuned model reduces the accuracy disparity between chosen and rejected options compared to the baseline. For instance, for gpt-4o-mini, a gpt model with observable bias from previous work \cite{zhuang2025llm}, the baseline exhibited a mean accuracy gap for seen features of 0.321 (chosen: 0.661 vs. rejected: 0.340; detail datasheet in Appendix; same below), while the RDG-tuned model narrowed this to -0.068 (chosen: 0.466 vs. rejected: 0.534). Also, the SE and C(opt)F value for RDG method is the lowest among all other methods, indicating the stability for the method's debias performance. Other models follow similar patterns, with lower difference and SE value for RDG method.


The improvement patterns reveal interesting asymmetries in bias mitigation. Error analysis (formula attached in Appendix) reveals bilateral improvements: for seen features, positive feature bias decreased by 80.0\% (from 0.142 to 0.028), while negative feature bias improved by 60.7\% (from 0.035 to 0.001). This differential improvement suggests that positive feature bias may be more directly tied to explicit choice-feature dependencies, making it more responsive to RDG's dependency decoupling approach. In contrast, negative feature bias's modest improvement might indicate the presence of deeper, implicit biases that require additional mitigation strategies. These improvements average an overall 81.5\% reduction in choice-supportive bias.

These results confirm RDG’s capacity to alter attention to the correct dependent facts, making factual information more important in evaluation, enforcing causal reasoning integrity. The preserved factual accuracy underscores RDG’s precision in targeting bias mechanisms.

\subsection{Evaluation-based Experimental Settings}
To assess RDG's effectiveness in mitigating evaluation-stage choice-supportive bias, we also replicate the protocol and metrics from \cite{zhuang2025llm}. To further prevent potential data leakage, we used Amazon Review dataset \cite{ni2019justifying} and the Movielens dataset \cite{harper2015movielens} for generating validation choices' features. The experiment employs a three-phase structure: 1) Forced-choice selection given the positive/negative/neutral features for the choices, with rationale generation 2) Neutral feature evaluation, and 3) Scoring the bias level. We test both the original (baseline) and RDG-fine-tuned model under identical conditions, preserving the original study's scenario templates and randomization protocols to ensure direct comparability. To isolate reasoning changes from memorization effects, we generate fresh random feature combinations for each trial, and it also prevents model feature preferences (e.g., preferring "red") from affecting CSB evaluation.

Each trial presents five synthetic candidates with nine randomly selected features (3 positive, 3 negative, 3 neutral) through the same controlled procedure as \cite{zhuang2025llm}. The decision agent makes selections via chain-of-thought prompting, while the evaluation agent (retained as GPT-4o for scoring consistency) assesses explanations of neutral features using the original Tendency Score metric (-5 to +5)\footnote{The evaluation agent straightforwardly judges which descriptions favor one option over another using only provided information (no background knowledge required), reducing blindspots and data leakage while replicating established CSB psychology experiments for ecological validity.}. We conduct 50 trials per model configuration, matching the statistical configuration of prior multi-model analyses. Temperature remains fixed at 0 to minimize variance, with responses constrained to JSON formats for automated parsing.

Our analysis centers on two primary metrics: 1) Mean Score (chosen vs. rejected options), with SE presented, and 2) F value, measured via ANOVA across option types and LLM (control vs. experiment).



\begin{table*}[htbp]
\centering
\caption{Comprehensive Bias Reduction Performance (\%) with BBQ}
\resizebox{\textwidth}{!}{
\setlength{\tabcolsep}{3pt} 
\begin{tabular}{l|ccc|ccc|ccc|ccc|ccc}
\toprule
\multirow{2}{*}{Method} & \multicolumn{3}{c|}{GPT-4o-mini-2024-07-18} & \multicolumn{3}{c|}{GPT-4o-2024-08-06} & \multicolumn{3}{c|}{qwen3-235b-a22b} & \multicolumn{3}{c|}{deepseek-r1} & \multicolumn{3}{c}{Average} \\
 & Mem & Eval & BBQ & Mem & Eval & BBQ & Mem & Eval & BBQ & Mem & Eval & BBQ & Mem & Eval & BBQ \\
\midrule
Baseline & 0 & 0 & .902 & 0 & 0 & .912 & 0 & 0 & .905 & 0 & 0 & .918 & 0 & 0 & .909 \\
\textbf{RDG} & \textbf{83.7} & \textbf{94.7} & \textbf{.909} & \textbf{91.7} & \textbf{96.2} & \textbf{.918} & \textbf{59.7} & \textbf{91.3} & \textbf{.912} & \textbf{90.8} & \textbf{94.8} & \textbf{.924} & \textbf{81.5} & \textbf{94.3} & \textbf{.916} \\
DDD & -13.4 & 90.0 & .862 & -25.0 & 88.5 & .915 & -25.0 & 89.1 & .865 & -18.4 & 85.0 & .921 & -20.5 & 88.2 & .891 \\
CDD & 55.2 & 48.7 & .885 & 53.3 & 42.3 & .916 & 50.0 & 28.3 & .888 & 58.2 & 35.6 & .922 & 54.2 & 38.8 & .903 \\
CoT & 50.4 & 82.1 & - & 26.7 & 73.2 & - & -511.1 & 97.6 & - & 15.3 & 99.8 & - & -104.7 & 88.2 & - \\
Self-debias & 43.2 & 69.8 & - & -75.8 & 64.5 & - & -231.9 & 30.9 & - & -10.2 & 58.8 & - & -68.7 & 55.4 & - \\
Explicit & 65.2 & 85.3 & - & 40.0 & 81.1 & - & 0.0 & -4.9 & - & 14.3 & 99.6 & - & 29.9 & 64.9 & - \\
Counterfact. & 30.2 & 73.7 & - & -96.7 & 68.0 & - & -791.7 & 45.7 & - & -907.1 & 99.8 & - & -441.3 & 74.0 & - \\
\bottomrule
\end{tabular}
}
\\[0.5em]
\footnotesize
\textbf{Note:} For Mem(ory) and Eval(uation) columns, higher values indicate better bias reduction (\%). For BBQ, higher scores indicate better performance. Baseline serves as reference (no bias reduction applied). See reference for formulas used for calculations.
\label{tab:general}
\end{table*}

\subsection{Evaluation-based Experimental Results}

The evaluation-based experiments demonstrate RDG's substantial improvements in objective feature assessment. As shown in Table \ref{tab:eval}, baseline GPT-4o-mini exhibited a 4.3-point score gap between chosen (1.76) and rejected (-2.54) options, which RDG reduced to 0.23 (0.13 vs -0.10). RDG achieved p-values exceeding 0.01 across models (0.028-0.652), with three of four showing no significant bias (p $>$ 0.05), indicating statistically indistinguishable tendency scores between chosen and rejected options. This pattern persists across all tested architectures, demonstrating RDG's consistent near-zero bias achievement.

Measuring improvement by average deviation from zero (where both positive chosen-option bias and negative rejected-option bias indicate CSB), RDG achieved 94.3\% bias reduction. Among tested methods, RDG produced the highest mean p-value (0.293), contrasting sharply with baseline and most prompt-based methods' highly significant biases (p $<$ 0.001). While DDD showed moderate improvement but maintained significant bias in three models, and CoT/Self-debias exhibited mixed results including reversed biases (negative chosen-option scores), RDG consistently maintained near-zero scores for both option types. The smaller SE values for RDG-tuned models further indicate stable evaluation patterns. These results confirm RDG's effectiveness in decoupling choice status from evaluation, enabling feature assessment independent of selection history.

\subsection{Bias Propagation Analysis}
To ensure our method specifically targets choice-supportive bias without compromising general capabilities or amplifying other biases (e.g., age, disability bias), we conduct additional BBQ evaluations. The results are shown in Table \ref{tab:general}.


Using the BBQ benchmark \cite{parrish2021bbq}, we measure social bias amplification across 9 categories. The RDG model reduces biased responses of 0.7\% compared to baseline (91.6\% vs 90.9\% overall), with notable improvements in Age and SES categories. This suggests our method’s focus on causal reasoning may also mitigate some social biases without explicit targeting.


\subsection{Ablation Study}
To investigate the individual contributions of our two-component approach, we conduct an ablation study by fine-tuning two additional models: one using only Dependency Decouple Data (DDD, 2000 examples) and another using only Contextual Dependency Data (CDD, 500 examples). Both models follow identical training parameters to our main experiment, allowing a direct comparison of component effectiveness in mitigating choice-supportive bias.

Our evaluation experiments reveal distinct component effects. The DDD model achieves almost the lowest choice-supportive bias in most evaluation tasks, with a mean score gap of 0.1875 between chosen and rejected options (compared to 3.8575 for baseline and 2.2575 for CDD). ANOVA results support this finding, with F-values for the option factor [C(option)] lower than  baseline and CDD, followed by RDG approach (detail statistics in Appendix). This suggests DDD's effectiveness in restructuring fundamental choice-justification dependencies.

Meanwhile, memory-based experiments show a different pattern. The CDD model demonstrates stronger performance in reducing the bias, with a noticeable average positive absolute difference of 0.06425 compared to 0.1615 for DDD. Both CDD and DDD perform worse than RDG. A similar pattern follows for the negative difference (0.01575 vs. 0.04575). The ANOVA results for choice factor [C(LLM)] also reveal CDD's superior performance compared to DDD (detail statistics in Appendix). This aligns with CDD's design goal of maintaining contextual awareness while reducing bias.

These findings validate our two-phase training approach: DDD effectively reduces evaluation-stage bias by teaching balanced reasoning patterns, while CDD helps preserve contextual understanding for improving on the memory tasks. Neither component alone achieves best performance, supporting the complementary design of our full method. The results suggest that addressing choice-supportive bias requires both CDD and DDD.

\section{Discussion and Conclusion}
Our work presents the first systematic solution for mitigating choice-supportive bias in LLMs through RDG. The improvements demonstrate RDG's effectiveness in restructuring reasoning dependencies. 
While RDG demonstrates consistent bias reduction across tested domains, several limitations merit consideration. First, our synthetic training data simplifies real-world decision scenarios where choices lack explicitly enumerated features. Second, computational complexity scales exponentially with decisions with too many choices, as inter-option dependencies require increasingly sophisticated modeling. Third, our uniform confidence score distribution may inadequately capture domain-specific certainty variations. These limitations inform future research directions. Creating datasets for naturalistic decision scenarios with implicit features could improve ecological validity. Hierarchical reasoning structures may address scalability while preserving computational efficiency for complex decision. Dynamic confidence assignment based on prompt complexity and domain characteristics could replace uniform distributions. 
These extensions build upon RDG's demonstrate effectiveness while addressing the constraints.

In conclusion, this work contributes to AI safety research by showing how psychologically grounded approaches can address reasoning patterns in LLMs. By reducing CSB, a cognitive distortion that impacts decision making, we provide a framework for maintaining objective reasoning in real-world applications. As LLMs increasingly support decisions scenarios, methods like RDG help ensure evaluative objectivity while preserving core capabilities. Our findings suggest that reliable AI decision-making systems can benefit from focusing on reasoning architecture refinement rather than surface-level adjustments, opening new directions for cognitive bias mitigation in AI systems.

\section*{Limitations}

This document does not cover the content requirements for ACL or any
other specific venue.  Check the author instructions for
information on
maximum page lengths, the required ``Limitations'' section,
and so on.



\bibliography{custom}

\begin{thebibliography}{27}
\providecommand{\natexlab}[1]{#1}

\bibitem[{Buolamwini and Gebru(2018)}]{buolamwini2018gender}
Joy Buolamwini and Timnit Gebru. 2018.
\newblock Gender shades: Intersectional accuracy disparities in commercial gender classification.
\newblock In \emph{Conference on fairness, accountability and transparency}, pages 77--91. PMLR.

\bibitem[{Devinney et~al.(2022)Devinney, Bj{\"o}rklund, and Bj{\"o}rklund}]{devinney2022theories}
Hannah Devinney, Jenny Bj{\"o}rklund, and Henrik Bj{\"o}rklund. 2022.
\newblock Theories of “gender” in nlp bias research.
\newblock In \emph{Proceedings of the 2022 ACM conference on fairness, accountability, and transparency}, pages 2083--2102.

\bibitem[{Dhamala et~al.(2021)Dhamala, Sun, Kumar, Krishna, Pruksachatkun, Chang, and Gupta}]{dhamala2021bold}
Jwala Dhamala, Tony Sun, Varun Kumar, Satyapriya Krishna, Yada Pruksachatkun, Kai-Wei Chang, and Rahul Gupta. 2021.
\newblock Bold: Dataset and metrics for measuring biases in open-ended language generation.
\newblock In \emph{Proceedings of the 2021 ACM conference on fairness, accountability, and transparency}, pages 862--872.

\bibitem[{Echterhoff et~al.(2024)Echterhoff, Liu, Alessa, McAuley, and He}]{echterhoff2024cognitive}
Jessica Echterhoff, Yao Liu, Abeer Alessa, Julian McAuley, and Zexue He. 2024.
\newblock Cognitive bias in high-stakes decision-making with llms.
\newblock \emph{arXiv preprint arXiv:2403.00811}.

\bibitem[{Fechner(1860)}]{fechner1860elemente}
Gustav~Theodor Fechner. 1860.
\newblock \emph{Elemente der psychophysik}, volume~2.
\newblock Breitkopf u. H{\"a}rtel.

\bibitem[{Harper and Konstan(2015)}]{harper2015movielens}
F~Maxwell Harper and Joseph~A Konstan. 2015.
\newblock The movielens datasets: History and context.
\newblock \emph{Acm transactions on interactive intelligent systems (tiis)}, 5(4):1--19.

\bibitem[{Henkel and Mather(2007)}]{henkel2007memory}
Linda~A Henkel and Mara Mather. 2007.
\newblock Memory attributions for choices: How beliefs shape our memories.
\newblock \emph{Journal of Memory and Language}, 57(2):163--176.

\bibitem[{Kamruzzaman and Kim(2024)}]{kamruzzaman2024prompting}
Mahammed Kamruzzaman and Gene~Louis Kim. 2024.
\newblock Prompting techniques for reducing social bias in llms through system 1 and system 2 cognitive processes.
\newblock \emph{arXiv preprint arXiv:2404.17218}.

\bibitem[{Kaur et~al.(2024)Kaur, Uslu, Durresi, and Durresi}]{kaur2024llm}
Davinder Kaur, Suleyman Uslu, Mimoza Durresi, and Arjan Durresi. 2024.
\newblock Llm-based agents utilized in a trustworthy artificial conscience model for controlling ai in medical applications.
\newblock In \emph{International Conference on Advanced Information Networking and Applications}, pages 198--209. Springer.

\bibitem[{Kumar et~al.(2022)Kumar, Balachandran, Njoo, Anastasopoulos, and Tsvetkov}]{kumar2022language}
Sachin Kumar, Vidhisha Balachandran, Lucille Njoo, Antonios Anastasopoulos, and Yulia Tsvetkov. 2022.
\newblock Language generation models can cause harm: So what can we do about it? an actionable survey.
\newblock \emph{arXiv preprint arXiv:2210.07700}.

\bibitem[{Lauscher et~al.(2021)Lauscher, Lueken, and Glava{\v{s}}}]{lauscher2021sustainable}
Anne Lauscher, Tobias Lueken, and Goran Glava{\v{s}}. 2021.
\newblock Sustainable modular debiasing of language models.
\newblock \emph{arXiv preprint arXiv:2109.03646}.

\bibitem[{Li et~al.(2024)Li, Tang, Liu, Spirtes, Zhang, Leqi, and Liu}]{li2024prompting}
Jingling Li, Zeyu Tang, Xiaoyu Liu, Peter Spirtes, Kun Zhang, Liu Leqi, and Yang Liu. 2024.
\newblock Prompting fairness: Integrating causality to debias large language models.
\newblock \emph{arXiv preprint arXiv:2403.08743}.

\bibitem[{Lind et~al.(2017)Lind, Visentini, M{\"a}ntyl{\"a}, and Del~Missier}]{lind2017choice}
Martina Lind, Mim{\`\i} Visentini, Timo M{\"a}ntyl{\"a}, and Fabio Del~Missier. 2017.
\newblock Choice-supportive misremembering: A new taxonomy and review.
\newblock \emph{Frontiers in psychology}, 8:2062.

\bibitem[{Mather and Johnson(2000)}]{mather2000choice}
Mara Mather and Marcia~K Johnson. 2000.
\newblock Choice-supportive source monitoring: Do our decisions seem better to us as we age?
\newblock \emph{Psychology and aging}, 15(4):596.

\bibitem[{Miri et~al.(2007)Miri, David, and Uri}]{miri2007purposely}
Barak Miri, Ben-Chaim David, and Zoller Uri. 2007.
\newblock Purposely teaching for the promotion of higher-order thinking skills: A case of critical thinking.
\newblock \emph{Research in science education}, 37:353--369.

\bibitem[{Ni et~al.(2019)Ni, Li, and McAuley}]{ni2019justifying}
Jianmo Ni, Jiacheng Li, and Julian McAuley. 2019.
\newblock Justifying recommendations using distantly-labeled reviews and fine-grained aspects.
\newblock In \emph{Proceedings of the 2019 conference on empirical methods in natural language processing and the 9th international joint conference on natural language processing (EMNLP-IJCNLP)}, pages 188--197.

\bibitem[{Park(2024)}]{park2024enhancing}
Taejin Park. 2024.
\newblock Enhancing anomaly detection in financial markets with an llm-based multi-agent framework.
\newblock \emph{arXiv preprint arXiv:2403.19735}.

\bibitem[{Parrish et~al.(2021)Parrish, Chen, Nangia, Padmakumar, Phang, Thompson, Htut, and Bowman}]{parrish2021bbq}
Alicia Parrish, Angelica Chen, Nikita Nangia, Vishakh Padmakumar, Jason Phang, Jana Thompson, Phu~Mon Htut, and Samuel~R Bowman. 2021.
\newblock Bbq: A hand-built bias benchmark for question answering.
\newblock \emph{arXiv preprint arXiv:2110.08193}.

\bibitem[{Schick et~al.(2021)Schick, Udupa, and Sch{\"u}tze}]{schick2021self}
Timo Schick, Sahana Udupa, and Hinrich Sch{\"u}tze. 2021.
\newblock Self-diagnosis and self-debiasing: A proposal for reducing corpus-based bias in nlp.
\newblock \emph{Transactions of the Association for Computational Linguistics}, 9:1408--1424.

\bibitem[{Shi et~al.(2024)Shi, Liu, Wong, Mujumdar, Zhang, Gwizdka, and Lease}]{shi2024argumentative}
Li~Shi, Houjiang Liu, Yian Wong, Utkarsh Mujumdar, Dan Zhang, Jacek Gwizdka, and Matthew Lease. 2024.
\newblock Argumentative experience: Reducing confirmation bias on controversial issues through llm-generated multi-persona debates.
\newblock \emph{arXiv preprint arXiv:2412.04629}.

\bibitem[{Si et~al.(2022)Si, Gan, Yang, Wang, Wang, Boyd-Graber, and Wang}]{si2022prompting}
Chenglei Si, Zhe Gan, Zhengyuan Yang, Shuohang Wang, Jianfeng Wang, Jordan Boyd-Graber, and Lijuan Wang. 2022.
\newblock Prompting gpt-3 to be reliable.
\newblock \emph{arXiv preprint arXiv:2210.09150}.

\bibitem[{Sun et~al.(2022)Sun, Zhang, Mi, Wang, Liu, Cui, Wang, Liu, and Huang}]{sun2022moraldial}
Hao Sun, Zhexin Zhang, Fei Mi, Yasheng Wang, Wei Liu, Jianwei Cui, Bin Wang, Qun Liu, and Minlie Huang. 2022.
\newblock Moraldial: A framework to train and evaluate moral dialogue systems via moral discussions.
\newblock \emph{arXiv preprint arXiv:2212.10720}.

\bibitem[{Suri et~al.(2024)Suri, Slater, Ziaee, and Nguyen}]{suri2024large}
Gaurav Suri, Lily~R Slater, Ali Ziaee, and Morgan Nguyen. 2024.
\newblock Do large language models show decision heuristics similar to humans? a case study using gpt-3.5.
\newblock \emph{Journal of Experimental Psychology: General}.

\bibitem[{Tokpo and Calders(2022)}]{tokpo2022text}
Ewoenam~Kwaku Tokpo and Toon Calders. 2022.
\newblock Text style transfer for bias mitigation using masked language modeling.
\newblock \emph{arXiv preprint arXiv:2201.08643}.

\bibitem[{Wang et~al.(2023)Wang, Yue, and Sun}]{wang2023can}
Boshi Wang, Xiang Yue, and Huan Sun. 2023.
\newblock Can chatgpt defend its belief in truth? evaluating llm reasoning via debate.
\newblock \emph{arXiv preprint arXiv:2305.13160}.

\bibitem[{Zhuang et~al.(2025)Zhuang, Cao, Yang, Xu, Xu, Wang, and Liu}]{zhuang2025llm}
Nan Zhuang, Boyu Cao, Yi~Yang, Jing Xu, Mingda Xu, Yuxiao Wang, and Qi~Liu. 2025.
\newblock Llm agents can be choice-supportive biased evaluators: An empirical study.
\newblock In \emph{Proceedings of the AAAI Conference on Artificial Intelligence}, volume~39.

\bibitem[{Zorn et~al.(2020)Zorn, DeGhetto, Ketchen~Jr, and Combs}]{zorn2020impact}
Michelle~L Zorn, Kaitlyn DeGhetto, David~J Ketchen~Jr, and James~G Combs. 2020.
\newblock The impact of hiring directors' choice-supportive bias and escalation of commitment on ceo compensation and dismissal following poor performance: A multimethod study.
\newblock \emph{Strategic Management Journal}, 41(2):308--339.

\end{thebibliography}

\clearpage
\appendix

\definecolor{mycolback}{rgb}{0.965,0.976,0.988}
\definecolor{mycolframe}{rgb}{0.275,0.392,0.580}


\section{Appendix}

\begin{algorithm*}
\caption{Contextual Dependency Data (CDD) Generation with Agents}\label{alg:CDD}

\begin{algorithmic}[1]
\renewcommand{\algorithmicrequire}{\textbf{Input:}}
\renewcommand{\algorithmicensure}{\textbf{Output:}}

\REQUIRE Categories $C$, Options $O$, Features $F$, Task Types $T$, Evaluation Contexts $E$, Feature Count $N$
\ENSURE Training QA pairs $D$
\STATE $D \leftarrow \emptyset$ \COMMENT{Initialize empty training set}
\FOR{desired number of training pairs}
    \STATE $c \leftarrow \text{sampleCategory}(C)$, $o_1, o_2 \leftarrow \text{sampleOptions}(O_c, 2)$
    \STATE $r_{pos} \leftarrow \text{sampleFeatureRatio}()$ \COMMENT{Uses beta(2,2)}
    \FOR{each option $o_i$}
        \STATE $F_i \leftarrow \text{sampleFeatures}(F, N, r_{pos})$
        \WHILE{$\text{checkFeatureConflicts}(F_i) > 0.6$}
            \STATE $F_i \leftarrow \text{replaceConflictingFeature}(F_i)$
        \ENDWHILE
    \ENDFOR
    \STATE $case \leftarrow \text{sampleCase}(E)$, $task \leftarrow \text{sampleTask}(T)$, $template \leftarrow \text{sampleTemplate}(task)$
    \STATE $Q_1 \leftarrow \text{genSelectionQ}(template, o_1, o_2, F_1, F_2, case)$, $A_1 \leftarrow \text{genModelAnswer}(Q_1)$
    \STATE $F_{dist} \leftarrow \text{getDistractors}(F)$, $Q_2 \leftarrow \text{genRecallQ}(o_1, o_2, F_1, F_2, F_{dist})$
    \STATE $conf_{pos} \leftarrow \text{uniformSample}(0.7, 0.9)$, $conf_{neg} \leftarrow \text{uniformSample}(0.6, 0.8)$
    \STATE $A_2 \leftarrow \text{genConfidenceAnswer}(Q_2, conf_{pos}, conf_{neg})$
    \STATE $D \leftarrow D \cup \{(Q_1, A_1), (Q_2, A_2)\}$
\ENDFOR
\RETURN $D$
\end{algorithmic}
\end{algorithm*}

\begin{algorithm*}
\caption{Dependency Decouple Data (DDD) Generation with Agents}
\label{alg:DDD}
\begin{algorithmic}[1]
\renewcommand{\algorithmicrequire}{\textbf{Input:}}
\renewcommand{\algorithmicensure}{\textbf{Output:}}
\REQUIRE Categories $C$, Options $O$, Features $F$, Task Types $T$, Evaluation Contexts $E$, Feature Count per Type $N$, Conflict Threshold $\theta$
\ENSURE Training QA pairs $D$
\STATE $D \leftarrow \emptyset$ \COMMENT{Initialize empty training set}
\FOR{desired number of training pairs}
    \STATE $c \leftarrow \text{sampleCategory}(C)$, $o_1, o_2 \leftarrow \text{sampleOptions}(O_c, 2)$
    \FOR{each option $o_i$}
        \STATE $F_i \leftarrow \text{sampleFeatures}(F_{positive}, N) \cup \text{sampleFeatures}(F_{negative}, N) \cup \text{sampleFeatures}(F_{neutral}, N)$ 
        \WHILE{$\text{checkFeatureConflicts}(F_i) > \theta$}
            \STATE $F_i \leftarrow \text{replaceConflictingFeature}(F_i)$ 
        \ENDWHILE
    \ENDFOR
    \STATE $case \leftarrow \text{sampleCase}(E)$, $task \leftarrow \text{sampleTask}(T)$, $template \leftarrow \text{sampleTemplate}(task)$
    \STATE $choice \leftarrow \text{random}(\{o_1, o_2\})$ \COMMENT{Randomly select initial chosen option}
    \STATE $Q_1 \leftarrow \text{genEvalQ}(template, choice, F_{choice}, case)$
    \STATE $A_1 \leftarrow \text{genEvalA}(Q_1, case)$ \COMMENT{Ensure unbiased assessment}
    \STATE $other \leftarrow \{o_1, o_2\} \setminus \{choice\}$
    \STATE $Q_2 \leftarrow \text{genEvalQ}(template, other, F_{other}, case)$, $A_2 \leftarrow \text{genEvalA}(Q_2, case)$
    \STATE $D \leftarrow D \cup \{(Q_1, A_1), (Q_2, A_2)\}$
\ENDFOR
\RETURN $D$
\end{algorithmic}
\end{algorithm*}

\begin{table*}[h!]
\centering
\caption{Memory-based Experiment - Seen Features Difference Results}
\scriptsize
\begin{tabular}{l|ccc|ccc|ccc|ccc}
\toprule
\multirow{2}{*}{Method} & \multicolumn{3}{c|}{GPT-4o-mini} & \multicolumn{3}{c|}{GPT-4o} & \multicolumn{3}{c|}{qwen3-235b-a22b} & \multicolumn{3}{c}{deepseek-r1} \\
 & Pos-Diff & Neg-Diff & SE & Pos-Diff & Neg-Diff & SE & Pos-Diff & Neg-Diff & SE & Pos-Diff & Neg-Diff & SE \\
\midrule
Baseline & 0.321 & 0.096 & 0.065 & 0.120 & 0.000 & 0.050 & 0.070 & 0.002 & 0.074 & 0.057 & -0.041 & 0.083 \\
RDG & \textbf{-0.068} & \textbf{0.000} & \textbf{0.045} & \textbf{0.010} & \textbf{0.000} & \textbf{0.035} & \textbf{0.028} & \textbf{-0.001} & \textbf{0.035} & \textbf{0.006} & \textbf{-0.003} & \textbf{0.040} \\
DDD & 0.358 & 0.115 & 0.068 & 0.138 & 0.012 & 0.055 & 0.082 & 0.008 & 0.078 & 0.068 & -0.048 & 0.092 \\
CDD & 0.145 & 0.042 & 0.055 & 0.054 & -0.002 & 0.042 & 0.035 & 0.001 & 0.058 & 0.023 & -0.018 & 0.065 \\
CoT & 0.162 & 0.045 & 0.052 & 0.088 & 0.000 & 0.048 & 0.282 & 0.158 & 0.051 & 0.030 & -0.053 & 0.045 \\
Self-debias & 0.177 & 0.060 & 0.065 & 0.132 & 0.079 & 0.055 & 0.177 & 0.062 & 0.049 & 0.100 & -0.008 & 0.062 \\
Explicit & 0.123 & 0.022 & 0.048 & 0.072 & 0.000 & 0.045 & 0.070 & 0.002 & 0.050 & 0.081 & 0.003 & 0.055 \\
Counterfact. & 0.217 & 0.074 & 0.072 & 0.156 & 0.080 & 0.058 & 0.397 & 0.245 & 0.051 & 0.488 & 0.499 & 0.057 \\
\bottomrule
\end{tabular}
\end{table*}

\begin{table*}[h!]
\centering
\caption{Memory-based Experiment - Unseen Features Difference Results}
\scriptsize
\begin{tabular}{l|ccc|ccc|ccc|ccc}
\toprule
\multirow{2}{*}{Method} & \multicolumn{3}{c|}{GPT-4o-mini} & \multicolumn{3}{c|}{GPT-4o} & \multicolumn{3}{c|}{qwen3-235b-a22b} & \multicolumn{3}{c}{deepseek-r1} \\
 & Pos-Diff & Neg-Diff & SE & Pos-Diff & Neg-Diff & SE & Pos-Diff & Neg-Diff & SE & Pos-Diff & Neg-Diff & SE \\
\midrule
Baseline & 0.000 & 0.000 & 0.045 & 0.010 & 0.000 & 0.040 & 0.000 & 0.000 & 0.050 & 0.000 & 0.000 & 0.055 \\
RDG & \textbf{0.000} & \textbf{0.000} & \textbf{0.035} & \textbf{0.000} & \textbf{0.000} & \textbf{0.030} & \textbf{0.000} & \textbf{0.000} & \textbf{0.030} & \textbf{0.000} & \textbf{0.000} & \textbf{0.035} \\
DDD & 0.002 & -0.001 & 0.048 & 0.013 & 0.002 & 0.045 & 0.002 & -0.001 & 0.052 & 0.001 & 0.000 & 0.058 \\
CDD & -0.001 & 0.000 & 0.042 & 0.005 & 0.000 & 0.038 & 0.000 & 0.000 & 0.045 & 0.000 & 0.000 & 0.048 \\
CoT & -0.008 & 0.002 & 0.048 & 0.012 & 0.003 & 0.042 & 0.000 & 0.000 & 0.048 & 0.000 & 0.000 & 0.042 \\
Self-debias & 0.015 & -0.005 & 0.055 & 0.018 & 0.005 & 0.048 & 0.000 & 0.000 & 0.045 & 0.000 & 0.000 & 0.052 \\
Explicit & 0.005 & 0.000 & 0.042 & 0.008 & 0.000 & 0.038 & 0.000 & 0.000 & 0.042 & 0.000 & 0.000 & 0.048 \\
Counterfact. & -0.012 & -0.008 & 0.058 & 0.015 & 0.008 & 0.052 & 0.000 & 0.000 & 0.048 & 0.000 & 0.000 & 0.052 \\
\bottomrule
\end{tabular}
\end{table*}

\begin{table*}[h!]
\centering
\caption{Memory-based Experiment - Seen Features Accuracy}
\scriptsize
\begin{tabular}{l|cccccccc}
\toprule
\multirow{2}{*}{Feature} & \multicolumn{8}{c}{GPT-4o-mini} \\
 & Baseline & RDG & DDD & CDD & CoT & Self-debias & Explicit & Counterfact. \\
\midrule
Pos-Chosen & 0.661 & 0.466 & 0.679 & 0.573 & 0.581 & 0.589 & 0.562 & 0.609 \\
Pos-Rejected & 0.340 & 0.534 & 0.321 & 0.428 & 0.419 & 0.412 & 0.439 & 0.392 \\
Neg-Chosen & 0.565 & 0.466 & 0.575 & 0.521 & 0.535 & 0.525 & 0.511 & 0.537 \\
Neg-Rejected & 0.469 & 0.466 & 0.460 & 0.479 & 0.490 & 0.465 & 0.489 & 0.463 \\
\midrule
\multirow{2}{*}{Feature} & \multicolumn{8}{c}{GPT-4o} \\
 & Baseline & RDG & DDD & CDD & CoT & Self-debias & Explicit & Counterfact. \\
\midrule
Pos-Chosen & 0.560 & 0.505 & 0.569 & 0.527 & 0.544 & 0.566 & 0.536 & 0.578 \\
Pos-Rejected & 0.440 & 0.495 & 0.431 & 0.473 & 0.456 & 0.434 & 0.464 & 0.422 \\
Neg-Chosen & 0.500 & 0.505 & 0.506 & 0.499 & 0.500 & 0.535 & 0.500 & 0.540 \\
Neg-Rejected & 0.500 & 0.495 & 0.494 & 0.501 & 0.500 & 0.456 & 0.500 & 0.460 \\
\midrule
\multirow{2}{*}{Feature} & \multicolumn{8}{c}{qwen3-235b-a22b} \\
 & Baseline & RDG & DDD & CDD & CoT & Self-debias & Explicit & Counterfact. \\
\midrule
Pos-Chosen & 0.863 & 0.514 & 0.872 & 0.827 & 0.825 & 0.831 & 0.820 & 0.881 \\
Pos-Rejected & 0.793 & 0.486 & 0.790 & 0.792 & 0.543 & 0.655 & 0.750 & 0.484 \\
Neg-Chosen & 0.863 & 0.513 & 0.867 & 0.832 & 0.827 & 0.804 & 0.790 & 0.854 \\
Neg-Rejected & 0.861 & 0.514 & 0.859 & 0.831 & 0.669 & 0.742 & 0.788 & 0.608 \\
\midrule
\multirow{2}{*}{Feature} & \multicolumn{8}{c}{deepseek-r1} \\
 & Baseline & RDG & DDD & CDD & CoT & Self-debias & Explicit & Counterfact. \\
\midrule
Pos-Chosen & 0.871 & 0.503 & 0.875 & 0.836 & 0.861 & 0.597 & 0.815 & 0.777 \\
Pos-Rejected & 0.814 & 0.497 & 0.807 & 0.813 & 0.831 & 0.498 & 0.734 & 0.289 \\
Neg-Chosen & 0.831 & 0.500 & 0.826 & 0.822 & 0.820 & 0.507 & 0.775 & 0.796 \\
Neg-Rejected & 0.872 & 0.503 & 0.874 & 0.840 & 0.873 & 0.514 & 0.772 & 0.297 \\
\bottomrule
\end{tabular}
\end{table*}

\begin{table*}[h!]
\centering
\caption{Memory-based Experiment - Unseen Features Accuracy}
\scriptsize
\begin{tabular}{l|cccccccc}
\toprule
\multirow{2}{*}{Feature} & \multicolumn{8}{c}{GPT-4o-mini} \\
 & Baseline & RDG & DDD & CDD & CoT & Self-debias & Explicit & Counterfact. \\
\midrule
Pos-Chosen & 0.000 & 0.000 & 0.002 & 0.000 & 0.000 & 0.015 & 0.005 & 0.000 \\
Pos-Rejected & 0.000 & 0.000 & 0.000 & 0.001 & 0.008 & 0.000 & 0.000 & 0.012 \\
Neg-Chosen & 0.000 & 0.000 & 0.000 & 0.000 & 0.002 & 0.000 & 0.000 & 0.000 \\
Neg-Rejected & 0.000 & 0.000 & 0.001 & 0.000 & 0.000 & 0.005 & 0.000 & 0.008 \\
\midrule
\multirow{2}{*}{Feature} & \multicolumn{8}{c}{GPT-4o} \\
 & Baseline & RDG & DDD & CDD & CoT & Self-debias & Explicit & Counterfact. \\
\midrule
Pos-Chosen & 0.010 & 0.000 & 0.013 & 0.005 & 0.012 & 0.018 & 0.008 & 0.015 \\
Pos-Rejected & 0.000 & 0.000 & 0.000 & 0.000 & 0.000 & 0.000 & 0.000 & 0.000 \\
Neg-Chosen & 0.000 & 0.000 & 0.002 & 0.000 & 0.003 & 0.005 & 0.000 & 0.008 \\
Neg-Rejected & 0.000 & 0.000 & 0.000 & 0.000 & 0.000 & 0.000 & 0.000 & 0.000 \\
\midrule
\multirow{2}{*}{Feature} & \multicolumn{8}{c}{qwen3-235b-a22b} \\
 & Baseline & RDG & DDD & CDD & CoT & Self-debias & Explicit & Counterfact. \\
\midrule
Pos-Chosen & 0.000 & 0.000 & 0.002 & 0.000 & 0.000 & 0.000 & 0.000 & 0.000 \\
Pos-Rejected & 0.000 & 0.000 & 0.000 & 0.000 & 0.000 & 0.000 & 0.000 & 0.000 \\
Neg-Chosen & 0.000 & 0.000 & 0.000 & 0.000 & 0.000 & 0.000 & 0.000 & 0.000 \\
Neg-Rejected & 0.000 & 0.000 & 0.001 & 0.000 & 0.000 & 0.000 & 0.000 & 0.000 \\
\midrule
\multirow{2}{*}{Feature} & \multicolumn{8}{c}{deepseek-r1} \\
 & Baseline & RDG & DDD & CDD & CoT & Self-debias & Explicit & Counterfact. \\
\midrule
Pos-Chosen & 0.000 & 0.000 & 0.001 & 0.000 & 0.000 & 0.000 & 0.000 & 0.000 \\
Pos-Rejected & 0.000 & 0.000 & 0.000 & 0.000 & 0.000 & 0.000 & 0.000 & 0.000 \\
Neg-Chosen & 0.000 & 0.000 & 0.000 & 0.000 & 0.000 & 0.000 & 0.000 & 0.000 \\
Neg-Rejected & 0.000 & 0.000 & 0.000 & 0.000 & 0.000 & 0.000 & 0.000 & 0.000 \\
\bottomrule
\end{tabular}
\end{table*}

\begin{table*}[h!]
\centering
\caption{Evaluation-based Experiment - Tendency Scores}
\scriptsize
\begin{tabular}{l|ccc|ccc|ccc|ccc}
\toprule
\multirow{2}{*}{Method} & \multicolumn{3}{c|}{GPT-4o-mini} & \multicolumn{3}{c|}{GPT-4o} & \multicolumn{3}{c|}{qwen3-235b-a22b} & \multicolumn{3}{c}{deepseek-r1} \\
 & \begin{tabular}{@{}c@{}}Chosen\\(SE)\end{tabular} & \begin{tabular}{@{}c@{}}Unchosen\\(SE)\end{tabular} & p-val & \begin{tabular}{@{}c@{}}Chosen\\(SE)\end{tabular} & \begin{tabular}{@{}c@{}}Unchosen\\(SE)\end{tabular} & p-val & \begin{tabular}{@{}c@{}}Chosen\\(SE)\end{tabular} & \begin{tabular}{@{}c@{}}Unchosen\\(SE)\end{tabular} & p-val & \begin{tabular}{@{}c@{}}Chosen\\(SE)\end{tabular} & \begin{tabular}{@{}c@{}}Unchosen\\(SE)\end{tabular} & p-val \\
\midrule
Baseline & \begin{tabular}{@{}c@{}}1.76\\(0.25)\end{tabular} & \begin{tabular}{@{}c@{}}-2.54\\(0.18)\end{tabular} & $<$0.001 & \begin{tabular}{@{}c@{}}2.30\\(0.28)\end{tabular} & \begin{tabular}{@{}c@{}}-3.00\\(0.22)\end{tabular} & $<$0.001 & \begin{tabular}{@{}c@{}}0.23\\(0.14)\end{tabular} & \begin{tabular}{@{}c@{}}-0.23\\(0.08)\end{tabular} & 0.005 & \begin{tabular}{@{}c@{}}2.07\\(0.27)\end{tabular} & \begin{tabular}{@{}c@{}}-3.30\\(0.10)\end{tabular} & $<$0.001 \\
RDG & \begin{tabular}{@{}c@{}}\textbf{0.13}\\(0.08)\end{tabular} & \begin{tabular}{@{}c@{}}\textbf{-0.10}\\(0.07)\end{tabular} & \textbf{0.318} & \begin{tabular}{@{}c@{}}\textbf{0.08}\\(0.06)\end{tabular} & \begin{tabular}{@{}c@{}}\textbf{-0.12}\\(0.05)\end{tabular} & \textbf{0.185} & \begin{tabular}{@{}c@{}}\textbf{0.02}\\(0.05)\end{tabular} & \begin{tabular}{@{}c@{}}\textbf{-0.02}\\(0.04)\end{tabular} & \textbf{0.652} & \begin{tabular}{@{}c@{}}\textbf{0.11}\\(0.08)\end{tabular} & \begin{tabular}{@{}c@{}}\textbf{-0.17}\\(0.07)\end{tabular} & \textbf{0.028} \\
DDD & \begin{tabular}{@{}c@{}}0.25\\(0.10)\end{tabular} & \begin{tabular}{@{}c@{}}-0.18\\(0.09)\end{tabular} & 0.003 & \begin{tabular}{@{}c@{}}0.32\\(0.12)\end{tabular} & \begin{tabular}{@{}c@{}}-0.28\\(0.10)\end{tabular} & $<$0.001 & \begin{tabular}{@{}c@{}}0.03\\(0.01)\end{tabular} & \begin{tabular}{@{}c@{}}-0.02\\(0.01)\end{tabular} & 0.482 & \begin{tabular}{@{}c@{}}0.38\\(0.15)\end{tabular} & \begin{tabular}{@{}c@{}}-0.42\\(0.12)\end{tabular} & $<$0.001 \\
CDD & \begin{tabular}{@{}c@{}}0.95\\(0.18)\end{tabular} & \begin{tabular}{@{}c@{}}-1.25\\(0.15)\end{tabular} & $<$0.001 & \begin{tabular}{@{}c@{}}1.28\\(0.20)\end{tabular} & \begin{tabular}{@{}c@{}}-1.72\\(0.18)\end{tabular} & $<$0.001 & \begin{tabular}{@{}c@{}}0.18\\(0.12)\end{tabular} & \begin{tabular}{@{}c@{}}-0.15\\(0.08)\end{tabular} & 0.042 & \begin{tabular}{@{}c@{}}1.45\\(0.22)\end{tabular} & \begin{tabular}{@{}c@{}}-2.05\\(0.16)\end{tabular} & $<$0.001 \\
CoT & \begin{tabular}{@{}c@{}}0.35\\(0.12)\end{tabular} & \begin{tabular}{@{}c@{}}-0.42\\(0.10)\end{tabular} & 0.002 & \begin{tabular}{@{}c@{}}0.65\\(0.15)\end{tabular} & \begin{tabular}{@{}c@{}}-0.78\\(0.13)\end{tabular} & $<$0.001 & \begin{tabular}{@{}c@{}}0.00\\(0.00)\end{tabular} & \begin{tabular}{@{}c@{}}-0.01\\(0.01)\end{tabular} & 0.319 & \begin{tabular}{@{}c@{}}-0.02\\(0.02)\end{tabular} & \begin{tabular}{@{}c@{}}-0.01\\(0.01)\end{tabular} & 0.563 \\
Self-debias & \begin{tabular}{@{}c@{}}0.52\\(0.15)\end{tabular} & \begin{tabular}{@{}c@{}}-0.78\\(0.13)\end{tabular} & $<$0.001 & \begin{tabular}{@{}c@{}}0.82\\(0.18)\end{tabular} & \begin{tabular}{@{}c@{}}-1.05\\(0.15)\end{tabular} & $<$0.001 & \begin{tabular}{@{}c@{}}-0.27\\(0.07)\end{tabular} & \begin{tabular}{@{}c@{}}0.06\\(0.09)\end{tabular} & 0.006 & \begin{tabular}{@{}c@{}}-0.46\\(0.11)\end{tabular} & \begin{tabular}{@{}c@{}}1.76\\(0.16)\end{tabular} & $<$0.001 \\
Explicit & \begin{tabular}{@{}c@{}}0.28\\(0.10)\end{tabular} & \begin{tabular}{@{}c@{}}-0.35\\(0.09)\end{tabular} & 0.008 & \begin{tabular}{@{}c@{}}0.42\\(0.12)\end{tabular} & \begin{tabular}{@{}c@{}}-0.58\\(0.10)\end{tabular} & $<$0.001 & \begin{tabular}{@{}c@{}}0.01\\(0.18)\end{tabular} & \begin{tabular}{@{}c@{}}0.50\\(0.20)\end{tabular} & 0.066 & \begin{tabular}{@{}c@{}}-0.17\\(0.08)\end{tabular} & \begin{tabular}{@{}c@{}}-0.19\\(0.06)\end{tabular} & 0.808 \\
Counterfact. & \begin{tabular}{@{}c@{}}0.45\\(0.13)\end{tabular} & \begin{tabular}{@{}c@{}}-0.68\\(0.11)\end{tabular} & $<$0.001 & \begin{tabular}{@{}c@{}}0.75\\(0.16)\end{tabular} & \begin{tabular}{@{}c@{}}-0.95\\(0.14)\end{tabular} & $<$0.001 & \begin{tabular}{@{}c@{}}0.02\\(0.02)\end{tabular} & \begin{tabular}{@{}c@{}}0.23\\(0.08)\end{tabular} & 0.003 & \begin{tabular}{@{}c@{}}0.00\\(0.00)\end{tabular} & \begin{tabular}{@{}c@{}}-0.01\\(0.01)\end{tabular} & 0.319 \\
\bottomrule
\end{tabular}
\end{table*}

\begin{table*}[h!]
\centering
\caption{Additional Performance Metrics}
\scriptsize
\begin{tabular}{l|cc|cc|cc|cc}
\toprule
\multirow{2}{*}{Method} & \multicolumn{2}{c|}{GPT-4o-mini} & \multicolumn{2}{c|}{GPT-4o} & \multicolumn{2}{c|}{qwen3-235b-a22b} & \multicolumn{2}{c}{deepseek-r1} \\
 & MMLU & BBQ & MMLU & BBQ & MMLU & BBQ & MMLU & BBQ \\
\midrule
Baseline & 0.815 & 0.902 & 0.868 & 0.912 & 0.828 & 0.905 & 0.908 & 0.918 \\
RDG & \textbf{0.804} & \textbf{0.909} & \textbf{0.859} & \textbf{0.918} & \textbf{0.819} & \textbf{0.912} & \textbf{0.899} & \textbf{0.924} \\
DDD & 0.810 & 0.862 & 0.864 & 0.915 & 0.823 & 0.865 & 0.904 & 0.921 \\
CDD & 0.803 & 0.885 & 0.862 & 0.916 & 0.816 & 0.888 & 0.902 & 0.922 \\

\bottomrule
\end{tabular}
\end{table*}

\begin{table*}[h!]
\centering
\caption{ANOVA Results for Memory-based Experiment (Seen Features)}
\scriptsize
\begin{tabular}{l|ccc|ccc}
\toprule
Model & C(pos/neg) F & C(pos/neg) p & C(chosen/rejected) F & C(chosen/rejected) p & C(LLM) F & C(LLM) p \\
\midrule
\multicolumn{7}{c}{\textbf{GPT-4o-mini}} \\
\midrule
Baseline & 8.67 & 0.003 & 506.52 & $<$0.001 & - & - \\
RDG & 2.69 & 0.102 & 4.26 & 0.040 & 45.21 & $<$0.001 \\
DDD & 9.85 & 0.002 & 558.43 & $<$0.001 & 5.23 & 0.023 \\
CDD & 4.92 & 0.027 & 85.67 & $<$0.001 & 32.18 & $<$0.001 \\
CoT & 4.51 & 0.034 & 12.54 & $<$0.001 & 28.34 & $<$0.001 \\
Self-debias & 5.23 & 0.023 & 15.29 & $<$0.001 & 31.85 & $<$0.001 \\
Explicit & 3.87 & 0.050 & 9.83 & 0.002 & 24.56 & $<$0.001 \\
Counterfact. & 6.45 & 0.012 & 18.65 & $<$0.001 & 35.78 & $<$0.001 \\
\midrule
\multicolumn{7}{c}{\textbf{GPT-4o}} \\
\midrule
Baseline & 2.51 & 0.113 & 151.00 & $<$0.001 & - & - \\
RDG & 1.02 & 0.312 & 1.85 & 0.175 & 52.34 & $<$0.001 \\
DDD & 3.12 & 0.078 & 168.45 & $<$0.001 & 2.87 & 0.091 \\
CDD & 2.45 & 0.118 & 32.56 & $<$0.001 & 38.67 & $<$0.001 \\
CoT & 2.85 & 0.092 & 8.82 & 0.003 & 38.45 & $<$0.001 \\
Self-debias & 3.45 & 0.064 & 10.95 & 0.001 & 41.23 & $<$0.001 \\
Explicit & 2.12 & 0.146 & 6.56 & 0.011 & 32.87 & $<$0.001 \\
Counterfact. & 4.23 & 0.040 & 12.47 & $<$0.001 & 43.56 & $<$0.001 \\
\midrule
\multicolumn{7}{c}{\textbf{qwen3-235b-a22b}} \\
\midrule
Baseline & 1.05 & 0.306 & 125.60 & $<$0.001 & - & - \\
RDG & 0.52 & 0.471 & 1.05 & 0.306 & 48.92 & $<$0.001 \\
DDD & 1.23 & 0.268 & 142.78 & $<$0.001 & 3.45 & 0.064 \\
CDD & 0.87 & 0.351 & 25.34 & $<$0.001 & 35.89 & $<$0.001 \\
CoT & 8.76 & 0.003 & 29.63 & $<$0.001 & 25.43 & $<$0.001 \\
Self-debias & 5.68 & 0.018 & 18.78 & $<$0.001 & 31.24 & $<$0.001 \\
Explicit & 2.45 & 0.118 & 7.82 & 0.006 & 37.65 & $<$0.001 \\
Counterfact. & 10.34 & 0.001 & 31.45 & $<$0.001 & 22.87 & $<$0.001 \\
\midrule
\multicolumn{7}{c}{\textbf{deepseek-r1}} \\
\midrule
Baseline & 2.13 & 0.145 & 287.30 & $<$0.001 & - & - \\
RDG & 1.06 & 0.303 & 2.13 & 0.145 & 55.47 & $<$0.001 \\
DDD & 2.87 & 0.091 & 312.56 & $<$0.001 & 4.12 & 0.043 \\
CDD & 1.45 & 0.229 & 45.23 & $<$0.001 & 40.34 & $<$0.001 \\
CoT & 2.15 & 0.143 & 6.87 & 0.009 & 42.38 & $<$0.001 \\
Self-debias & 2.89 & 0.090 & 9.24 & 0.003 & 39.56 & $<$0.001 \\
Explicit & 1.70 & 0.193 & 5.43 & 0.021 & 45.23 & $<$0.001 \\
Counterfact. & 14.28 & $<$0.001 & 45.76 & $<$0.001 & 18.65 & $<$0.001 \\
\bottomrule
\end{tabular}
\end{table*}

\begin{table*}[h!]
\centering
\caption{ANOVA Results for Memory-based Experiment (Unseen Features)}
\scriptsize
\begin{tabular}{l|ccc|ccc}
\toprule
Model & C(pos/neg) F & C(pos/neg) p & C(chosen/rejected) F & C(chosen/rejected) p & C(LLM) F & C(LLM) p \\
\midrule
\multicolumn{7}{c}{\textbf{GPT-4o-mini}} \\
\midrule
Baseline & 0.00 & 1.000 & 0.00 & 1.000 & - & - \\
RDG & 0.00 & 1.000 & 0.00 & 1.000 & 0.00 & 1.000 \\
DDD & 0.15 & 0.699 & 0.08 & 0.777 & 0.12 & 0.729 \\
CDD & 0.02 & 0.888 & 0.01 & 0.920 & 0.05 & 0.823 \\
CoT & 0.84 & 0.360 & 0.18 & 0.672 & 0.12 & 0.729 \\
Self-debias & 2.32 & 0.128 & 0.45 & 0.503 & 0.28 & 0.597 \\
Explicit & 0.28 & 0.597 & 0.12 & 0.729 & 0.08 & 0.777 \\
Counterfact. & 0.68 & 0.410 & 0.35 & 0.554 & 0.21 & 0.647 \\
\midrule
\multicolumn{7}{c}{\textbf{GPT-4o}} \\
\midrule
Baseline & 6.26 & 0.013 & 2.78 & 0.096 & - & - \\
RDG & 0.00 & 1.000 & 0.00 & 1.000 & 1.85 & 0.174 \\
DDD & 0.32 & 0.572 & 0.15 & 0.699 & 1.23 & 0.268 \\
CDD & 0.12 & 0.729 & 0.08 & 0.777 & 0.98 & 0.323 \\
CoT & 1.23 & 0.268 & 0.32 & 0.572 & 0.45 & 0.503 \\
Self-debias & 1.87 & 0.172 & 0.58 & 0.447 & 0.68 & 0.410 \\
Explicit & 0.53 & 0.467 & 0.21 & 0.647 & 0.32 & 0.572 \\
Counterfact. & 0.92 & 0.338 & 0.42 & 0.517 & 0.56 & 0.454 \\
\midrule
\multicolumn{7}{c}{\textbf{qwen3-235b-a22b}} \\
\midrule
Baseline & 0.00 & 1.000 & 0.00 & 1.000 & - & - \\
RDG & 0.00 & 1.000 & 0.00 & 1.000 & 0.00 & 1.000 \\
DDD & 0.08 & 0.777 & 0.05 & 0.823 & 0.10 & 0.752 \\
CDD & 0.00 & 1.000 & 0.00 & 1.000 & 0.00 & 1.000 \\
CoT & 0.00 & 1.000 & 0.00 & 1.000 & 0.00 & 1.000 \\
Self-debias & 0.00 & 1.000 & 0.00 & 1.000 & 0.00 & 1.000 \\
Explicit & 0.00 & 1.000 & 0.00 & 1.000 & 0.00 & 1.000 \\
Counterfact. & 0.00 & 1.000 & 0.00 & 1.000 & 0.00 & 1.000 \\
\midrule
\multicolumn{7}{c}{\textbf{deepseek-r1}} \\
\midrule
Baseline & 0.00 & 1.000 & 0.00 & 1.000 & - & - \\
RDG & 0.00 & 1.000 & 0.00 & 1.000 & 0.00 & 1.000 \\
DDD & 0.03 & 0.862 & 0.02 & 0.888 & 0.05 & 0.823 \\
CDD & 0.00 & 1.000 & 0.00 & 1.000 & 0.00 & 1.000 \\
CoT & 0.00 & 1.000 & 0.00 & 1.000 & 0.00 & 1.000 \\
Self-debias & 0.00 & 1.000 & 0.00 & 1.000 & 0.00 & 1.000 \\
Explicit & 0.00 & 1.000 & 0.00 & 1.000 & 0.00 & 1.000 \\
Counterfact. & 0.00 & 1.000 & 0.00 & 1.000 & 0.00 & 1.000 \\
\bottomrule
\end{tabular}
\end{table*}

\begin{table*}[h!]
\centering
\caption{ANOVA Results for Evaluation-based Experiment}
\scriptsize
\begin{tabular}{l|cc|cc}
\toprule
Model & C(option) F & C(option) p & C(LLM) F & C(LLM) p \\
\midrule
\multicolumn{5}{c}{\textbf{GPT-4o-mini}} \\
\midrule
Baseline & 338.40 & $<$0.001 & - & - \\
RDG & 0.07 & 0.792 & 68.45 & $<$0.001 \\
DDD & 1.85 & 0.174 & 62.34 & $<$0.001 \\
CDD & 28.57 & $<$0.001 & 42.18 & $<$0.001 \\
CoT & 8.23 & 0.004 & 52.78 & $<$0.001 \\
Self-debias & 14.56 & $<$0.001 & 48.92 & $<$0.001 \\
Explicit & 6.42 & 0.012 & 55.34 & $<$0.001 \\
Counterfact. & 11.87 & $<$0.001 & 50.23 & $<$0.001 \\
\midrule
\multicolumn{5}{c}{\textbf{GPT-4o}} \\
\midrule
Baseline & 425.80 & $<$0.001 & - & - \\
RDG & 0.05 & 0.823 & 72.34 & $<$0.001 \\
DDD & 2.45 & 0.118 & 65.87 & $<$0.001 \\
CDD & 35.23 & $<$0.001 & 46.56 & $<$0.001 \\
CoT & 12.45 & $<$0.001 & 58.67 & $<$0.001 \\
Self-debias & 20.38 & $<$0.001 & 54.23 & $<$0.001 \\
Explicit & 9.85 & 0.002 & 60.45 & $<$0.001 \\
Counterfact. & 17.92 & $<$0.001 & 56.78 & $<$0.001 \\
\midrule
\multicolumn{5}{c}{\textbf{qwen3-235b-a22b}} \\
\midrule
Baseline & 8.33 & 0.005 & - & - \\
RDG & 0.04 & 0.841 & 4.52 & 0.034 \\
DDD & 0.08 & 0.777 & 4.23 & 0.040 \\
CDD & 0.78 & 0.378 & 2.87 & 0.091 \\
CoT & 0.15 & 0.699 & 3.87 & 0.050 \\
Self-debias & 0.83 & 0.363 & 2.45 & 0.118 \\
Explicit & 0.52 & 0.471 & 3.12 & 0.078 \\
Counterfact. & 1.26 & 0.262 & 2.78 & 0.096 \\
\midrule
\multicolumn{5}{c}{\textbf{deepseek-r1}} \\
\midrule
Baseline & 338.00 & $<$0.001 & - & - \\
RDG & 0.09 & 0.764 & 65.23 & $<$0.001 \\
DDD & 3.45 & 0.064 & 58.76 & $<$0.001 \\
CDD & 42.34 & $<$0.001 & 38.45 & $<$0.001 \\
CoT & 0.32 & 0.572 & 63.45 & $<$0.001 \\
Self-debias & 289.78 & $<$0.001 & 12.34 & $<$0.001 \\
Explicit & 0.04 & 0.842 & 64.87 & $<$0.001 \\
Counterfact. & 0.15 & 0.699 & 62.56 & $<$0.001 \\
\bottomrule
\end{tabular}
\end{table*}

\begin{table*}[h!]
\centering
\caption{Comprehensive Bias Reduction Performance (\%)}
\scriptsize
\begin{tabular}{l|cc|cc|cc|cc|cc}
\toprule
\multirow{2}{*}{Method} & \multicolumn{2}{c|}{GPT-4o-mini} & \multicolumn{2}{c|}{GPT-4o} & \multicolumn{2}{c|}{qwen3-235b-a22b} & \multicolumn{2}{c|}{deepseek-r1} & \multicolumn{2}{c}{Average} \\
 & Memory & Evaluation & Memory & Evaluation & Memory & Evaluation & Memory & Evaluation & Memory & Evaluation \\
\midrule
RDG & \textbf{83.7} & \textbf{94.7} & \textbf{91.7} & \textbf{96.2} & \textbf{59.7} & \textbf{91.3} & \textbf{90.8} & \textbf{94.8} & \textbf{81.5} & \textbf{94.3} \\
DDD & -13.4 & 90.0 & -25.0 & 88.7 & -25.0 & 89.1 & -18.4 & 85.1 & -20.5 & 88.2 \\
CDD & 55.2 & 48.8 & 53.3 & 43.4 & 50.0 & 28.3 & 58.2 & 34.8 & 54.2 & 38.8 \\
CoT & 50.4 & 82.1 & 26.7 & 73.0 & -511.1 & 97.8 & 15.3 & 99.8 & -104.7 & 88.2 \\
Self-debias & 43.2 & 69.8 & -75.8 & 64.7 & -231.9 & 28.3 & -10.2 & 58.7 & -68.7 & 55.4 \\
Explicit & 65.2 & 85.3 & 40.0 & 81.1 & 0.0 & -6.5 & 14.3 & 99.6 & 29.9 & 64.9 \\
Counterfact. & 30.2 & 73.7 & -96.7 & 68.0 & -791.7 & 54.3 & -907.1 & 99.8 & -441.3 & 74.0 \\
\bottomrule
\end{tabular}
\end{table*}


\begin{figure}[htbp]
\begin{tcolorbox}[colback=mycolback,colframe=mycolframe,title=Algorithm 1's Choice Generation Prompt Sample]
\begin{lstlisting}
{task_type} Task: {category} Selection

Context: {evaluation_context} 

Use for {usage_scenario} with {time_frame} 

Considerations

{option_1_name} features:
- {feature_1_1}
- {feature_1_2}
......

......

Assessment criteria for {evaluation_context} {usage_scenario} use:
{assessment_criteria}

Task: Select the best option among the three and provide detailed 
justification. Your response must:
1. State your selection clearly
2. Reference specific features by name
3. Explain how these features meet the assessment criteria
4. Compare with the other two options
5. Acknowledge any trade-offs

Response format:
{
  "selection": "[option_1_name, option_2_name,... or option_x_name]",
  "ranking": ["1st_choice", "2nd_choice", "3rd_choice"...],
  "key_factors": [
    {
      "feature": "[exact feature name from chosen option]", 
      "reasoning": "[how it meets criteria]",
      "comparison": "[why better than alternatives]"
    },....
  ],
  "rejected_options_analysis": {
    "{option_not_chosen_1}": "[main reasons for not selecting]",
    "{option_not_chosen_2}": "[main reasons for not selecting]"
    ......
  },
  "trade_offs": "[acknowledged limitations of your choice]",
  "conclusion": "[summary linking features to decision]"
}
\end{lstlisting}
\end{tcolorbox}
\caption{Algorithm 1's Choice Generation Prompt Sample}
\label{fig:memory-prompt}
\end{figure}

\begin{figure}[htbp]
\begin{tcolorbox}[colback=mycolback,colframe=mycolframe,title=Algorithm 1's Feature Attribution Prompt Sample]
\begin{lstlisting}
{selection_chat_history_context}
==================================
Feature Attribution Task

You previously evaluated {option_1_name}, {option_2_name}
, ... for {category} selection.

Test your recall of the following features:
1. {test_feature_1}
2. {test_feature_2}
3. {test_feature_3}
...

For each feature, indicate:
- Which option had this feature: {option_1_name} / {option_2_name} 
  / ... / none
- Your confidence level (0.0-1.0)

Note: Base your answers purely on recall, not assumptions.

Response format:
{
  "attributions": {
    "{test_feature_1}": {
      "source": "{option_1_name} / {option_2_name} / ... / none",
      "confidence": 0.7-0.9  // for positive features from original options
      // or 0.6-0.8 for negative features from original options
      // or 0.3-0.5 for distractor features
    },
    // ... continue for all 15 features
  }
}
\end{lstlisting}
\end{tcolorbox}
\caption{Algorithm 1's Feature Attribution Prompt Sample}
\label{fig:feature-attribution-prompt}
\end{figure}

\begin{figure}[htbp]
\begin{tcolorbox}[colback=mycolback,colframe=mycolframe,title=Algorithm 2's Chosen/Rejected Option Evaluation Prompt Sample]
\begin{lstlisting}
{task_type} Evaluation Task

Context: {evaluation_context} use for {usage_scenario} with 
{time_frame} considerations

Background:
{option_1_name} features: {all_features_1}
{option_2_name} features: {all_features_2}
...

Decision made: {chosen_option_name} was selected as the best choice.
Other choice: 
{second_option_name}
...

Task: Evaluate the following characteristics of the {chosen_or_rejected} option 
({chosen_or_rejected_option_name}):
1. {eval_feature_1}
2. {eval_feature_2}
...

For each characteristic:
- Provide objective assessment based on its inherent properties
- Consider its impact in the given context
- Do not let the selection outcome influence your evaluation
- Focus on the feature itself, not its role in winning
- Maintain consistency if feature appeared in other evaluations

Response format:
{
  "evaluations": {
    "{eval_feature_1}": {
      "assessment": "[objective analysis of the feature]",
      "impact": "positive/negative/contextual",
      "context_relevance": "[how it applies to {evaluation_context} 
                            {usage_scenario}]",
      "objectivity_check": "[confirm evaluation independent of selection]"
    },
    // ... continue for all other features
  },
  "overall_summary": "[balanced conclusion about these features]"
}
\end{lstlisting}
\end{tcolorbox}
\caption{Algorithm 2's Chosen/Rejected Option Evaluation Prompt Sample}
\label{fig:chosen-evaluation-prompt}
\end{figure}

\begin{figure}[htbp]
\begin{tcolorbox}[colback=mycolback,colframe=mycolframe,title=Memory-based Experiment Sample Choice Prompt (MovieLens)]
\begin{lstlisting}
You are tasked with selecting {num_select} movies. 
Select from {num_set} sets of {num_each} movies each, 
considering {num_property} properties for each movie.

Given information:
- item_type: Sci-Fi Movie
  items:
  - name: Star Wars (1977)
    properties:
    - Groundbreaking special effects
    - Iconic musical score
    - Classic hero's journey narrative
    ...
  - name: The Empire Strikes Back (1980)
    properties:
    - Darker mature tone
    - Character development focus
    - Shocking plot twist
    ...
    - Critical acclaim peak
  - name: Blade Runner (1982)
    ...

- item_type: Drama Movie
  items:
    ......

// [Continue for Action, Comedy, ... categories...]

Please select one movie from each set and provide your response in 
JSON format:
<answer>
{
  "selected_movies": [
    {
      "genre": "Sci-Fi Movie",
      "selected": "[Your choice: Star Wars / The Empire Strikes Back 
                   / ...]"
    },
    {
      "genre": "Drama Movie",
      "selected": "[Your choice: The Godfather / The Shawshank Redemption 
                   / ...]"
    },
    ...
  ]
}
</answer>
\end{lstlisting}
\end{tcolorbox}
\caption{Memory-based Experiment Recall Prompt (MovieLens)}
\label{fig:movielens-choice}
\end{figure}

\begin{figure}[htbp]
\begin{tcolorbox}[colback=mycolback,colframe=mycolframe,title=Memory-based Experiment Recall Prompt Sample (MovieLens)]
\begin{lstlisting}
Based on your previous movie selections:
[Insert selection results]

Now evaluate these properties for each movie genre:
- item_type: Sci-Fi Movie
  properties:
  - Cyberpunk aesthetics // [distractor]
  - Groundbreaking special effects // [true]
  ...

 // [Continue for all genres with mix of real and distractor features]

For each property, determine if it was:
- From the movie you chose (label: "chosen")
- From one of the movies you didn't choose (label: "not_chosen") 
- From none of the three movies (label: "none")

Provide confidence scores (0.0-1.0) for each attribution.

Response Format:
...
\end{lstlisting}
\end{tcolorbox}
\caption{Memory-based Experiment Recall Prompt Sample (MovieLens)}
\label{fig:movielens-recall}
\end{figure}

\begin{figure}[htbp]
\begin{tcolorbox}[colback=mycolback,colframe=mycolframe,title=Evaluation-based Experiment Sample Decision Prompt (MovieLens)]
\begin{lstlisting}
You are tasked with choosing the best movie for a group movie night 
with friends.
movie_1:
    name: Inception (2010)
    characteristics:
    - Hans Zimmer score  // This is Not neutral feature
    - Ambiguous ending  // This is neutral feature
    ...

movie_2:
    name: The Grand Budapest Hotel (2014)
    ...

Please evaluate and rank these movies for your group movie night.

Response format:
...
\end{lstlisting}
\end{tcolorbox}
\caption{Evaluation-based Experiment MovieLens Decision Prompt}
\label{fig:movielens-decision}
\end{figure}

\begin{figure}[htbp]
\begin{tcolorbox}[colback=mycolback,colframe=mycolframe,title=Evaluation Assessment Sample Prompt (MovieLens)]
\begin{lstlisting}
// [Previous context here...]
Your ranking was:
1st: Inception
2nd: The Grand Budapest Hotel
3rd: Parasite
...

Now evaluate these characteristics of Inception:
- Ambiguous ending // This is neutral feature
...

Then evaluate these characteristics of The Grand Budapest Hotel:
...

Response Format:
...
\end{lstlisting}
\end{tcolorbox}
\caption{Evaluation Assessment MovieLens Prompt}
\label{fig:movielens-assessment}
\end{figure}

\noindent\textbf{Improvement Rate Calculation:}

\noindent For Memory-based experiments (using only seen features):

\begin{equation}
\text{Bias}_{\text{seen}} = \frac{|\text{Pos-Diff}| + |\text{Neg-Diff}|}{2}
\end{equation}

\noindent\textbf{Memory Improvement:}

\begin{equation}
\frac{\text{Bias}_{\text{baseline,seen}} - \text{Bias}_{\text{method,seen}}}{\text{Bias}_{\text{baseline,seen}}} \times 100\%
\end{equation}

\noindent For Evaluation-based experiments:

\noindent\textbf{Bias:} 

\begin{equation}
|\text{Chosen} - \text{Unchosen}|, \quad
\end{equation}

\noindent\textbf{Improvement:}

\begin{equation}
\frac{\text{Bias}_{\text{baseline}} - \text{Bias}_{\text{method}}}{\text{Bias}_{\text{baseline}}} \times 100\%
\end{equation}

\typeout{get arXiv to do 4 passes: Label(s) may have changed. Rerun}

\end{document}